\documentclass{article}

\usepackage{PRIMEarxiv}

\usepackage[utf8]{inputenc} 
\usepackage[T1]{fontenc}    
\usepackage{hyperref}       
\usepackage{url}            
\usepackage{booktabs}       
\usepackage{amsfonts}       
\usepackage{nicefrac}       
\usepackage{microtype}      
\usepackage{lipsum}
\usepackage{fancyhdr}       
\usepackage{graphicx}       
\graphicspath{{media/}}     
\usepackage{amsmath, xparse}
\usepackage{xcolor}
\usepackage{color}
\usepackage{colortbl}

\usepackage{array}
\usepackage{multirow}
\PassOptionsToPackage{normalem}{ulem}
\usepackage{ulem}
\providecommand{\tabularnewline}{\\}

\title{Open-DDVM: A Reproduction and Extension of Diffusion Model for Optical Flow Estimation}

\author{
  Qiaole Dong  \\
  School of Data Science, Fudan University \\
  \texttt{qldong18@fudan.edu.cn} \\
  \And
  Bo Zhao \\
  Beijing Academy of Artificial Intelligence  \\
  \texttt{zhaobo@baai.ac.cn} \\
  \AND
  Yanwei Fu \\
  School of Data Science, Fudan University \\
  \texttt{yanweifu@fudan.edu.cn} \\
}

\begin{document}
\maketitle

\begin{abstract}

Recently, Google proposes DDVM~\cite{saxena2023surprising} which for the first time demonstrates that a general diffusion model for image-to-image translation task works impressively well on optical flow estimation task without any specific designs like RAFT~\cite{teed2020raft}. However, DDVM is still a closed-source model with the expensive and private Palette-style pretraining. In this technical report, we present the first open-source DDVM by reproducing it. We study several design choices and find those important ones. By training on 40k public data with 4 GPUs, our reproduction achieves comparable performance to the closed-source DDVM. The code and model have been released in \url{https://github.com/DQiaole/FlowDiffusion_pytorch}.
\end{abstract}


\section{Introduction}
Optical flow is a fundamental modality in computer vision, which provides key information for various real-world applications, \emph{e.g.}, video frame interpolation~\cite{jiang2018super}, video inpainting~\cite{gao2020flow}, and action recognition~\cite{sun2018optical}. Optical flow estimation aims to estimate a per-pixel displacement vector field between two consecutive video frames. Previous optical flow estimation methods~\cite{jiang2021learning,zheng2022dip,sun2022skflow,dong2023rethinking,huang2022flowformer} have achieved remarkable progress based on the successful architecture of Recurrent All-Pairs Field Transforms (RAFT)~\cite{teed2020raft}, which iteratively updates a flow field with GRU-based operator. Recently, denoising diffusion probabilistic models~\cite{ho2020denoising} have emerged as novel method for distribution modelling. Saxen et al. for the first time apply diffusion model for optical flow estimation and achieve impressively good performance as introduced in DDVM~\cite{saxena2023surprising}.

DDVM~\cite{saxena2023surprising} takes a noisy version of the optical flow as input and concatenates two RGB images as the condition. It employs the Efficient U-Net~\cite{saharia2022photorealistic} architecture to predict the noise-free estimation of the 2-channel optical flow, with the training loss penalizing residual error in the denoised map. Besides, to train a generic diffusion model for optical flow estimation without task-specific inductive biases, DDVM utilizes massively synthetic datasets, including AutoFlow~\cite{sun2021autoflow}, FlyingThings3D~\cite{mayer2016large}, TartanAir~\cite{wang2020tartanair}, and Kubric~\cite{greff2022kubric}. Additionally, DDVM requires a Palette-style~\cite{saharia2022palette} pretraining on huge image datasets. During inference, it employs a strategy called coarse-to-fine for high-resolution input, which refines the noised estimation in a patch-wise manner based on coarse results from resized low-resolution images. For more details about DDVM, please refer to the original paper~\cite{saxena2023surprising}.

\begin{figure*}
\centering
\includegraphics[width=0.85\linewidth]{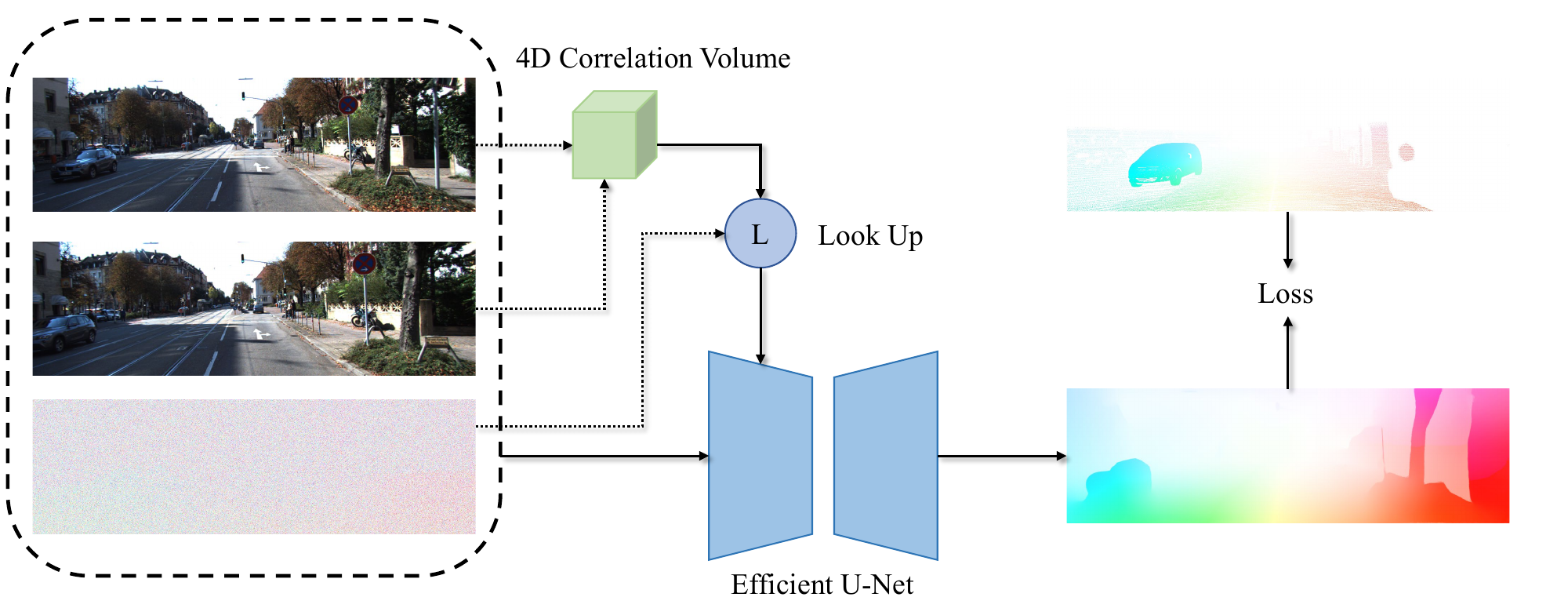}
\caption{Overview of our correlation-volume assisted diffusion model for optical flow estimation.\label{fig:overview}}
\end{figure*}

As mentioned above, training a generic diffusion model for optical flow estimation without task-specific inductive biases necessitates massive synthetic datasets and a longer training process. In this paper, we make a preliminary study and propose to take advantage of multi-scale correlation volume as a prior for optical flow estimation inspired by classic work~\cite{teed2020raft}. As shown in Fig.~\ref{fig:overview}, we first extract the feature for two input images on 1/8 resolution by a simple ResNet. We then construct a 4-layer multi-scale correlation volume through matrix multiplication and average pooling. We also perform lookups operation defined in RAFT~\cite{teed2020raft} on the 4D correlation volume with the current noised flow, which results in a single feature map containing all local correlation information. Our U-Net accepts both 8-channel tensors (two 3-channel RGB images and one 2-channel noised flow) and the correlation feature as input. Specifically, the U-Net encodes the 8-channel tensor and concatenates with the correlation feature map at 1/8 resolution of the U-Net encoder. The denoised optical flow is predicted through a decoder part network. We train the whole U-Net similarly as in DDVM~\cite{saxena2023surprising}.

\section{Experiments}

\paragraph{Dataset and Implementation Details.} We train our model on AutoFlow~\cite{sun2021autoflow} dataset of 320x448 resolution, which consists of 40k synthetic image pairs and ground-truth optical flow. Following previous works~\cite{teed2020raft, saxena2023surprising}, we evaluate the zero-shot performance of our model on the synthetic Sintel~\cite{butler2012naturalistic} and real-world KITTI~\cite{geiger2012we} datasets. We adopt the Efficient U-Net architecture~\cite{saharia2022photorealistic} as our backbone network, without the Palette-style pretraining. The batch-size is 64. AdamW is utilized as the default optimizer. We also enable the exponential-moving-average (EMA) during model training. We train the denoising U-Net with DDPM model~\cite{ho2020denoising} of 64 steps on four NVIDIA A800 GPUs. Besides, to ease the memory requirement and speed up the training process, we utilize DeepSpeed~\cite{rasley2020deepspeed} and mixed precision training with data type \textit{bf16}~\cite{teich2018tearing}. We initialize our learning rate with $1\times 10^{-4}$ and half the learning rate when encountering an unstable training stage due to mixed precision training.

\subsection{Ablation Study}
In this subsection, we determine several important factors for training the diffusion model for optical flow estimation.

\paragraph{Normalization.} As shown in the left part of Tab.~\ref{tab:ablation}, normalizing the input and output into the range of $[-1,1]$ using the height and width of input can greatly improve the zero-shot performance. Besides, we also observe that normalization helps stabilize the training process with more smoother loss curve.

\paragraph{Augmentation.} We compare the data augmentation operations from RAFT~\cite{teed2020raft}, including photometric and geometric augmentation, with the one from RAFT-it~\cite{sun2022disentangling}. We find that utilizing data augmentation from RAFT-it~\cite{sun2022disentangling}, which is also used in DDVM~\cite{saxena2023surprising}, can boost the performance, as shown in Tab.~\ref{tab:ablation}.

\paragraph{Noise Augmentation.} We further study the Gaussian noise augmentation in this experiment. Prior work like PWC-Net~\cite{sun2019models} observed that training without  Gaussian noise can improve the performance of PWC-Net, while recent RAFT-it~\cite{sun2022disentangling}, which focuses on improving the training process of optical flow estimation models, employs the Gaussian noise augmentation. In our experiment shown in Tab.~\ref{tab:ablation}, we find that Gaussian noise augmentation causes little difference. However, we find that adding noise can stabilize the model training. Thus, we follow the augmentation from RAFT-it in our later experiments.

\begin{figure*}
\centering
\includegraphics[width=0.85\linewidth]{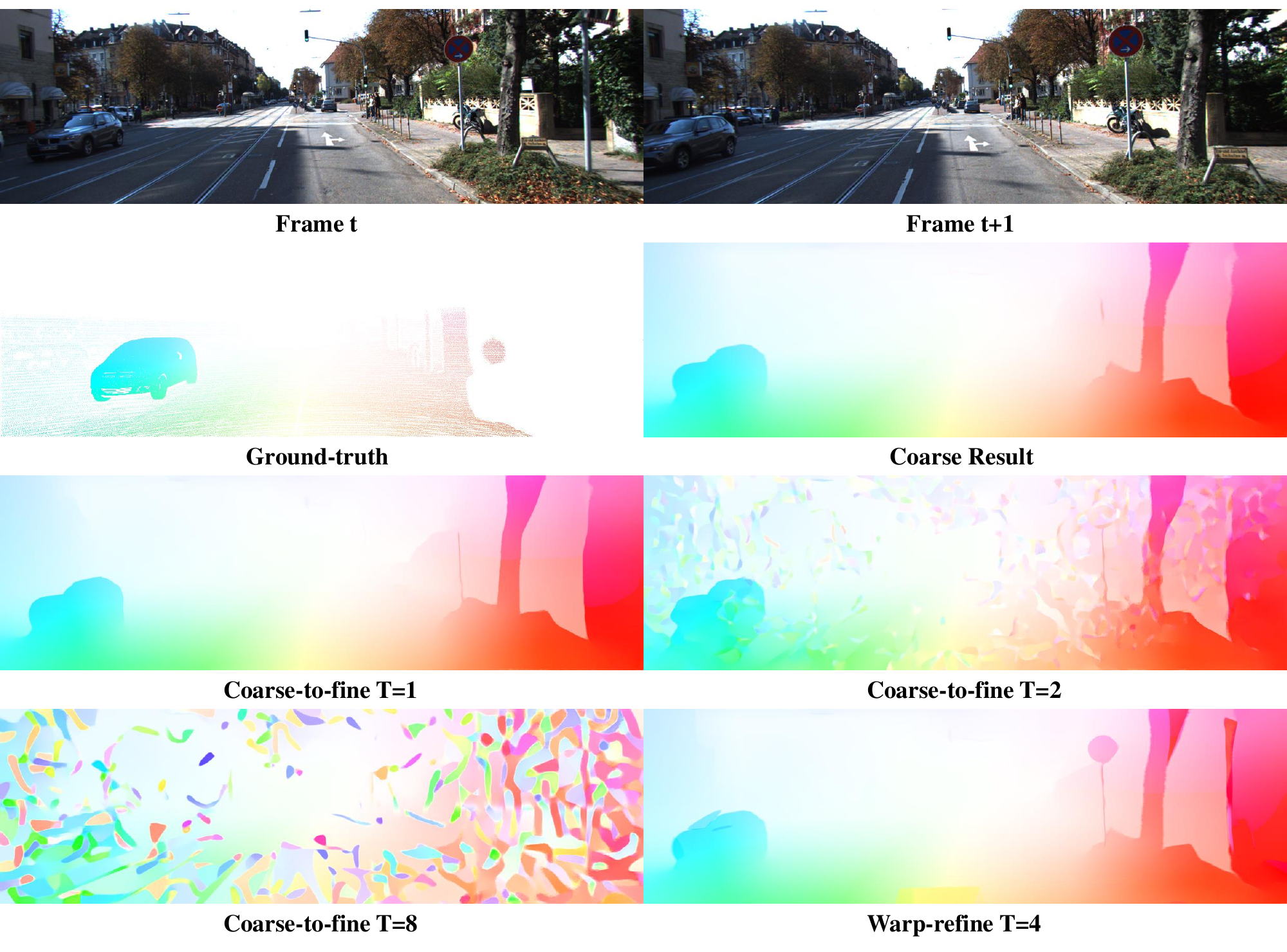}
\caption{Qualitative comparison with Coarse-to-fine refinement~\cite{saxena2023surprising} on KITTI-15. Our proposed Warp-refine performs better in maintaining the whole coarse structure and refining the fine details. However, we find that Coarse-to-fine refinement with 2 steps already introduces inconsistent noise into the final optical flow. Here $T$ represents the total steps used during the refinement stage.\label{fig:c2f_qualitative}}
\end{figure*}

\paragraph{Refinement.} DDVM~\cite{saxena2023surprising} performs inference in a coarse-to-fine manner. It first estimates the flow over the entire field of view at low resolution. It then up-samples the flow to the original resolution as $f$ and runs forward diffusion to an intermediate time step $T$, getting a noised flow. Finally, it denoises the flow in a patch-wise manner and merges the patches. However, we find that setting $T\geq 2$ will result in local inconsistency in the final denoised flow, as shown in Fig.~\ref{fig:c2f_qualitative}. Setting $T=1$ can improve the performance as expected, shown in Tab.~\ref{tab:ablation} and Fig.~\ref{fig:c2f_qualitative}. Therefore, we propose another refinement method, called warp-refine. Specifically, we first backward the second image with up-sampled coarse flow. We then estimate the residual flow between the warped image and the first image in a patch-wise manner with a small amount of time steps. As shown in Tab.~\ref{tab:ablation} and Fig.~\ref{fig:c2f_qualitative}, our warp-refine can improve the performance and capture more fine details.

\begin{table}
 \small
\centering
\caption{\emph{Left}: Ablation studies on AutoFlow~\cite{sun2021autoflow}. We evaluate the models on the downsampled KITTI-15 training set of 320x448 resolution and report the end-point-error (EPE). \emph{Right}: Ablation studies of refinement on the KITTI-15 training set with model trained on AutoFlow. The time step indicates the number of steps used during the refinement stage. The parameters used in our final model are
underlined.\label{tab:ablation}}
    \begin{tabular}{m{0.5\textwidth}m{0.5\textwidth}}
    \begin{tabular}{ccc}
\toprule
Experiment & Method & KITTI-15\tabularnewline
\midrule
\tabularnewline
\multicolumn{3}{c}{\noindent\emph{Training: 20k on AutoFlow}}\tabularnewline
\hline
\multirow{2}{*}{Normalize Flow} & No & 15.8\tabularnewline
 & \uline{Yes} & 7.4\tabularnewline
\hline
\tabularnewline
\multicolumn{3}{c}{\noindent\emph{Training: 100k on AutoFlow}}\tabularnewline
\hline
\multirow{2}{*}{Augmentation} & RAFT~\cite{teed2020raft} & 5.2\tabularnewline
 & \uline{RAFT-it}~\cite{sun2022disentangling} & 3.4\tabularnewline
\hline
\multirow{2}{*}{Noise Augmentation} & No & 3.5\tabularnewline
 & \uline{Yes} & 3.4\tabularnewline
\bottomrule
\end{tabular}
        &
        \begin{tabular}{cccc}
\toprule
 & Time Step & EPE & Fl-all\tabularnewline
\midrule
Without & - & 5.57 & 20.22\%\tabularnewline
\hline
\multirow{2}{*}{Coarse-to-fine~\cite{saxena2023surprising}} & T = 8 & 27.60 & 58.65\%\tabularnewline
 & \uline{T=1} & \textbf{5.53} & \textbf{19.67\%}\tabularnewline
\hline
\multirow{2}{*}{Warp-refine} & T=8 & 5.54 & \textbf{18.54\%}\tabularnewline
 & \uline{T=4} & \textbf{5.44} & 18.57\%\tabularnewline
\bottomrule
\end{tabular}
        \\
    \end{tabular}
\end{table}

\subsection{Optical Flow Estimation Results}
In this section, we report the zero-shot optical flow estimation performance on Sintel and KITTI-15 when trained on AutoFlow. We also report the results of DDVM with Palette-style pretraining as in Tab.~\ref{tab:main}. Our reproduced model trained from the scratch with 900k iterations on AutoFlow results in an EPE of 3.76 in Sintel Final pass and a Fl-all of 18.57\% in KITTI-15, as shown in the third line of Tab.~\ref{tab:main}. Besides, our modified network based on a simple correlation volume as introduced in Fig.~\ref{fig:overview} indeed performs better than the original network with the same iteration. With only 1/3 iterations, our correlation volume based diffusion model can even approach the performance of the original diffusion model with 900k iterations. We believe that introducing more sophisticated inductive biases into simple U-Net can further improve the performance and speed up the training of the diffusion model for optical flow estimation in the future.

\begin{table}
\setlength{\tabcolsep}{3pt}
\small
\centering
\caption{Zero-shot optical flow estimation results on Sintel and KITTI-15 training set. All models are trained on AutoFlow~\cite{sun2021autoflow}.\label{tab:main}}
\begin{tabular}{cccccccc}
\toprule
 & Pretraining & Network & Iteration & Sintel.clean & Sintel.final & KITTI EPE & KITTI Fl-all\tabularnewline
\midrule
DDVM~\cite{saxena2023surprising} & Palette-style & Efficient U-Net & Unknow & 2.04 & 2.55 & 4.47 & 16.59\%\tabularnewline
Ours & From the Scratch & Efficient U-Net & 305k & 2.96 & 3.97 & 6.21 & 20.38\%\tabularnewline
Ours & From the Scratch & Efficient U-Net & 900k & 2.77 & 3.76 & 5.44 & 18.57\%\tabularnewline
Ours & From the Scratch & Efficient U-Net + Corr. Volume & 305k & 2.98 & 3.85 & 5.53 & 19.04\%\tabularnewline
\bottomrule
\end{tabular}
\end{table}

\section{Conclusion}
In this technical report, we reproduce the DDVM and contribute the first open-source diffusion model for optical flow estimation. With limited resources, our reproduced DDVM achieves comparable performance to the original one. Our experiments reveal that some specific designs, such as employing correlation volume from prior optical flow models, are important for accurate optical flow estimation. We hope this open-source model can advance the research of optical flow estimation.

{
    \small
    \bibliographystyle{plain}
    \bibliography{references}
}

\end{document}